\documentclass{article}
\usepackage[utf8]{inputenc}
\usepackage[T1]{fontenc}
\usepackage{times}
\usepackage{helvet}
\usepackage{courier}

\usepackage{amsmath}
\usepackage{amssymb}
\usepackage{amsfonts}

\usepackage{booktabs}
\usepackage{multirow}
\usepackage{makecell}
\usepackage{array}
\usepackage{tabularx}

\usepackage{graphicx}
\usepackage{xcolor}
\usepackage{caption}

\usepackage{algorithm}
\usepackage{algpseudocode}

\usepackage{enumitem}
\usepackage{tcolorbox}
\usepackage{float}
\usepackage{nicefrac}
\usepackage{microtype}

\usepackage{url}
\usepackage[colorlinks=true, linkcolor=blue, citecolor=blue, urlcolor=blue]{hyperref}

\usepackage{tikz}

\usepackage[numbers,square]{natbib}

\usepackage{arxiv}

\title{SwiftRepertoire: Few-Shot Immune-Signature Synthesis via Dynamic Kernel Codes}

\author{
    Rong Fu \\
    Independent Researcher \\
    Corresponding author \and
    Yang Li \\
    Independent Researcher \and
    Yabin Jin \\
    Independent Researcher \and
    Jiekai Wu \\
    Independent Researcher \and
    Chunlei Meng \\
    Independent Researcher \and
     Simon Fong \\
    Independent Researcher \and
}

\renewcommand{\headeright}{}
\renewcommand{\undertitle}{}
\renewcommand{\shorttitle}{SwiftRepertoire}

\hypersetup{
pdftitle={SwiftRepertoire: Few-Shot Immune-Signature Synthesis via Dynamic Kernel Codes},
pdfsubject={q-bio.QM, q-bio.GN},
pdfauthor={Rong Fu, Yang Li, Yabin Jin, Jiekai Wu, Chunlei Meng,Simon Fong},
pdfkeywords={T-cell receptor repertoire, Few-shot learning, Immune signature, Cancer detection, Meta-learning, Adapter synthesis},
}

\begin{document}
\maketitle

\begin{abstract}
Repertoire-level analysis of T cell receptors offers a biologically grounded signal for disease detection and immune monitoring, yet practical deployment is impeded by label sparsity, cohort heterogeneity, and the computational burden of adapting large encoders to new tasks. We introduce a framework that synthesizes compact task-specific parameterizations from a learned dictionary of prototypes conditioned on lightweight task descriptors derived from repertoire probes and pooled embedding statistics. This synthesis produces small adapter modules applied to a frozen pretrained backbone, enabling immediate adaptation to novel tasks with only a handful of support examples and without full model fine-tuning. The architecture preserves interpretability through motif-aware probes and a calibrated motif discovery pipeline that links predictive decisions to sequence-level signals. Together, these components yield a practical, sample-efficient, and interpretable pathway for translating repertoire-informed models into diverse clinical and research settings where labeled data are scarce and computational resources are constrained. 
\end{abstract}


\section{Introduction}

Early detection of malignancy remains central to improving patient outcomes, and conventional modalities such as organ-specific imaging and circulating biomarkers have shaped current screening practice ~\cite{li2020t, sivapalan2023liquid, huang2024liquid}. Recent advances in high-throughput immunogenomics have highlighted the immune repertoire as a complementary window into neoplastic processes because tumor-reactive T cells and their receptors carry signals of antigen exposure and immune selection ~\cite{beshnova2020novo, pan2022t, cai2024deep}. The variable regions of T cell receptors, and in particular the complementarity determining region three, encode sequence motifs that mediate peptide recognition and can reflect tumor-associated perturbations ~\cite{qian2024deeplion2, xu2022deeplion, sidhom2021deeptcr}. These biological observations motivate repertoire-based approaches for diagnosis and monitoring that do not depend on tumor somatic markers or limited biomarker panels. Computational methods for repertoire analysis have evolved from statistical enrichment tests and handcrafted feature models to deep learning systems that exploit sequence context and repertoire structure ~\cite{smith2021identification, zhang2024berttcr, cai2024deep}. Multiple-instance formulations and attention-based pooling have been proposed to aggregate information across repertoires ~\cite{kim2022multiple, shao2021transmil, sidhom2021deeptcr}. At the same time, protein language models and self-supervised encoders have offered transferable representations that improve downstream binding and classification performance ~\cite{dounas2024learning, leem2022deciphering}. Despite these advances, several persistent challenges limit practical adoption. First, many repertoire tasks confront severe label sparsity and a long-tail distribution of antigen or disease labels, which causes conventional supervised learners to generalize poorly to rare or unseen tasks ~\cite{gao2023pan, wang2023meta}. Second, adapting large pre-trained encoders to new tasks using full fine-tuning is computationally costly and may overfit when only a few labeled examples exist. Third, parameter-efficient adaptation techniques have been developed, yet there remains a need for methods that can produce immediate, interpretable, and statistically calibrated task-specific behaviors from minimal supervision ~\cite{zhao2023prototype, lv2024hyperlora, wu2023pacia}. Finally, clinical translation benefits from models that not only predict but also provide motif-level interpretability and calibrated statistical signals for downstream decision making.

To address these gaps we propose SwiftRepertoire, a conditional fast-weight framework that synthesizes compact adapter parameters via prototype-based retrieval conditioned on succinct task descriptors. The method learns a dictionary of geometry-aware prototypes and a constrained retrieval mechanism that enforces sparsity during parameter synthesis. Task descriptors are derived from lightweight repertoire probes and pooled embedding statistics to capture the salient task-specific geometry without heavy computation. The resulting fast weights are applied to a frozen encoder via small adapter modules, enabling instantaneous adaptation to a new task using only a handful of support examples. SwiftRepertoire couples this adaptation pathway with a motif-aware interpretability pipeline that performs calibrated motif testing on synthesized attention probes, yielding interpretable sequence signals alongside predictive outputs.

Our contributions are as follows. We introduce a retrieval-driven fast-weight synthesis procedure that produces sparse, rank-aware adapters compatible with large pretrained encoders. We design geometry-preserving prototype canonicalization and a proximal solver that enforces hard sparsity in the retrieval process. We propose a compact task descriptor built from pooled embedding gradients and repertoire probes to enable low-overhead task conditioning. We integrate a nested calibration workflow for motif-level interpretability that links predictive adaptation with statistically controlled sequence discovery. Together, these elements create a practical, sample-efficient, and interpretable few-shot adaptation system tailored for repertoire-level diagnostics.
\begin{figure*}[t]
  \centering
  \includegraphics[width=0.88\textwidth]{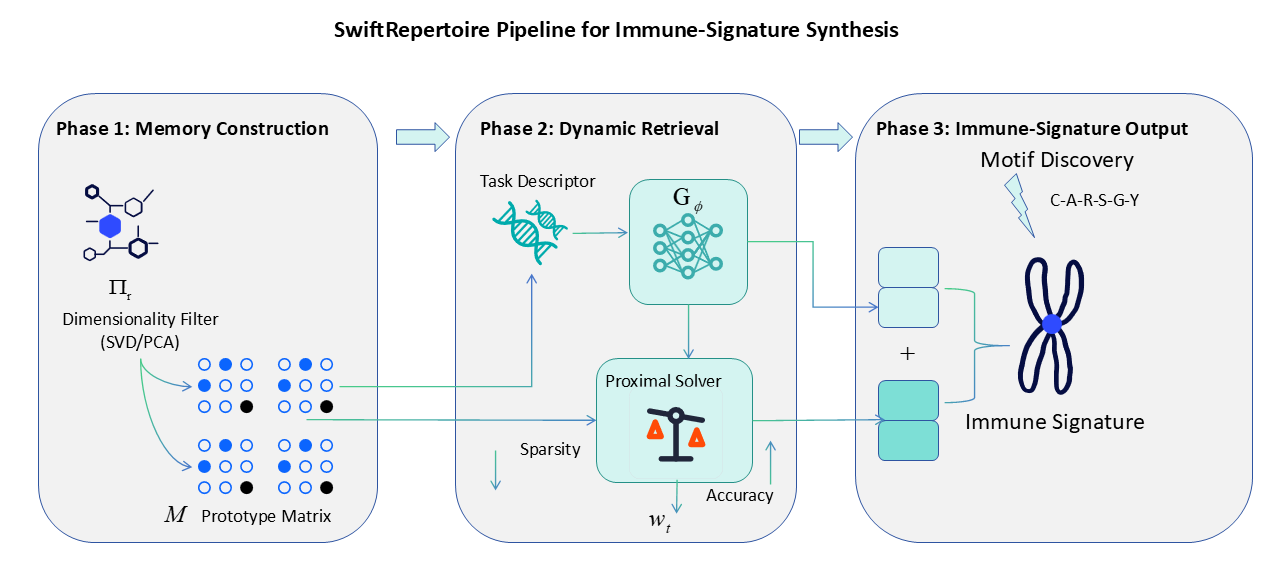}
  \caption[Overview of the SwiftRepertoire pipeline]{%
    \textbf{Overview of the SwiftRepertoire pipeline.}
    The figure summarizes the main stages and diagnostics used in SwiftRepertoire:
    low-dimensional adapter diagnostics (PCA energy and Fisher summaries), formal Fisher-energy hypothesis testing (bootstrap + Bonferroni correction), prototype construction and canonicalization, and the constrained proximal retrieval training with nested anti-leakage partitions and diagnostics.
    Neural-ODE components with stiff-stable integrators are integrated into the retrieval and descriptor pipelines to provide continuous-time modeling where appropriate.
  }
  \label{fig:metatcr-overview}
\end{figure*}

\section{Related Work}

\subsection{Repertoire biology and diagnostic motivation}
T cell receptor repertoires record antigen-driven selection and contain signals that are clinically useful for cancer detection and immune monitoring. Prior studies report that repertoire-derived summary statistics and clonotype-level features can discriminate cancer from non-cancer cohorts ~\cite{beshnova2020novo} and reveal spatial or temporal heterogeneity in tumour-infiltrating lymphocytes ~\cite{joshi2019spatial}. These biological observations motivate computational pipelines that extract CDR3 motifs and clonotype patterns as complementary diagnostics to imaging and circulating tumour biomarkers ~\cite{li2025circulating, pan2022t}.

\subsection{Representations and deep sequence models}
A broad spectrum of sequence-based models has been applied to TCR analysis, including convolutional encoders, transformer and BERT-style architectures, and multiple-instance formulations that pool repertoire evidence. Self-supervised protein language models provide contextual embeddings which, when fine-tuned on downstream tasks, typically yield substantial performance gains ~\cite{xu2022deeplion, zhang2024berttcr}. Recent work demonstrates that contrastive, cross-epitope fine-tuning aligns TCR and peptide embeddings in a shared latent space and improves epitope ranking and binding prediction relative to sequence-similarity baselines ~\cite{im2025sequence}. Complementary studies show that pretrained PLMs can also reveal confounding biophysical patterns such as viral immune mimicry, highlighting the importance of pairing representational improvements with interpretability and mechanistic analysis ~\cite{ofer2025protein, qian2024deeplion2}. These advances motivate using fine-tuned protein LMs both for pairwise binding prediction and for repertoire-level classification ~\cite{chen2023contiformer}.

\subsection{Few-shot learning and task adaptation}
The long-tailed distribution of peptide labels and the limited availability of annotated repertoires have driven meta-learning and few-shot approaches for TCR tasks. Memory-augmented meta-learners and task-conditioned modules borrow strength across tasks to generalize to rare or unseen peptides, and pan-peptide meta-learning frameworks have demonstrated robustness to novel antigens in clinically relevant applications ~\cite{gao2023pan}. Other methods emphasize rapid adaptation mechanisms that avoid full fine-tuning while maintaining strong performance from a handful of labeled examples ~\cite{wang2023meta, yue2025meta}.

\subsection{Parameter-efficient adaptation, prototypes, and interpretability}
Practical deployment demands methods that trade adaptation capacity against computation and memory. Parameter-efficient adapters, prototype retrieval and fast-weight schemes compress per-task knowledge into compact atoms and enable sparse retrieval to synthesize task-specific parameters with modest overhead ~\cite{zhao2023prototype, lv2024hyperlora}. At the same time, interpretable motif discovery and statistically calibrated testing remain essential for translational use; coupling sparse adaptation with motif-level inference yields models that are both predictive and biologically interpretable ~\cite{zhang2022tip, wu2023pacia}. Prototype-driven schemes have been combined with motif analysis to provide localized explanations of model decisions ~\cite{qian2024deeplion2, pan2022t}.

\subsection{Positioning}
Representation learning, multiple-instance learning and meta-learning form a strong methodological foundation, but practical challenges persist: principled criteria for adapter rank selection, prototype designs that preserve geometry, sparsity-aware retrieval integrated with calibrated inference, and an end-to-end pipeline that merges few-shot adaptation with interpretable motif discovery. SwiftRepertoire addresses these gaps by constructing geometry-preserving prototypes, synthesizing sparse, rank-aware adapters conditioned on compact task descriptors, and embedding adaptation within a nested motif calibration workflow to preserve statistical rigor and biological insight.

\begin{figure}[h]
\centering
\includegraphics[width=0.8\columnwidth]{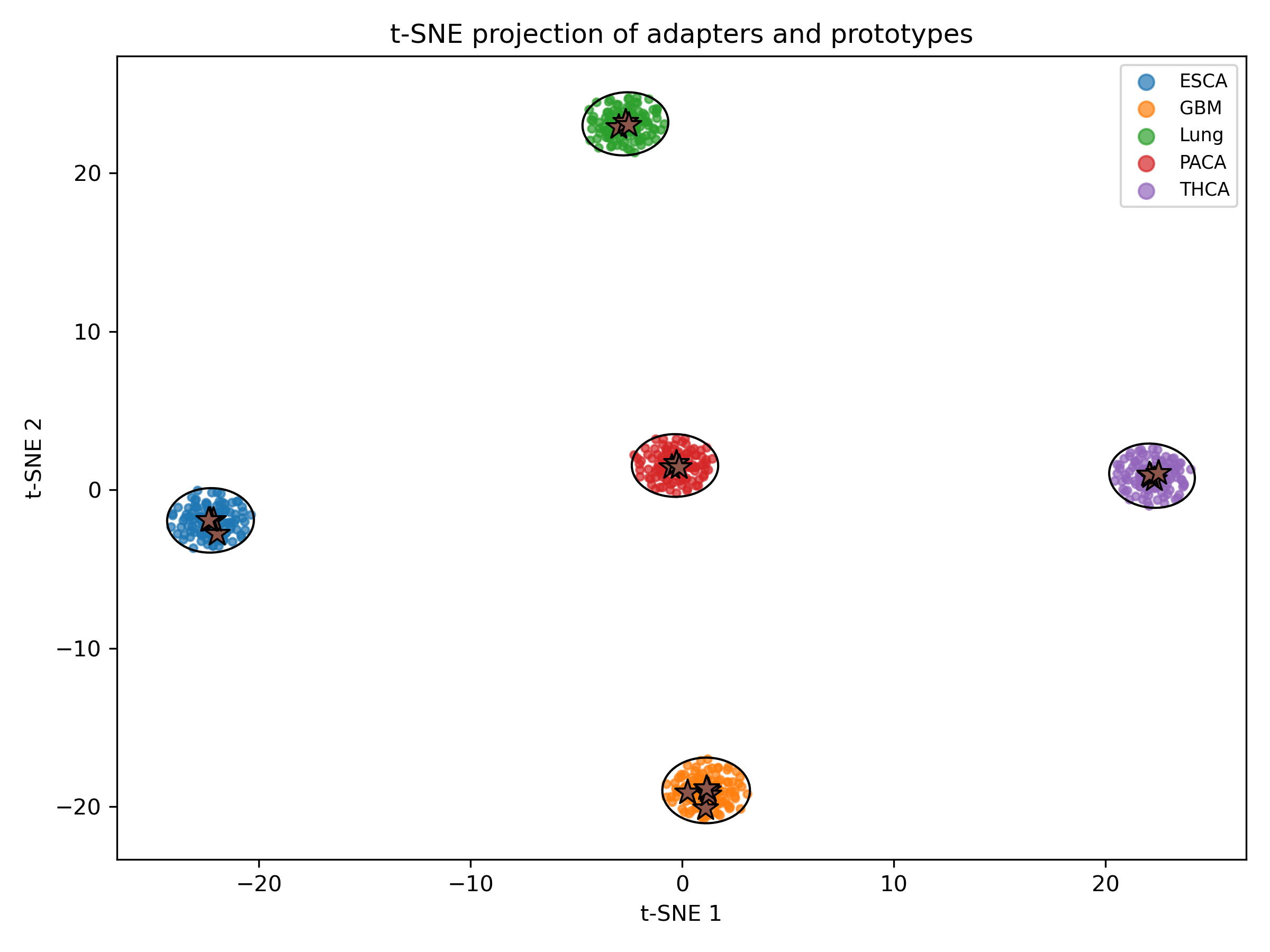}
\caption{t-SNE projection of adapter vectors and learned prototypes. Points are colored by disease / task ID; black ellipses highlight major clusters and black stars mark prototype centroids. This visualization verifies whether adapters exhibit task-separable low-dimensional structure.}
\label{fig:tsne_projection}
\end{figure}

\begin{figure}[h]
\centering
\includegraphics[width=0.8\columnwidth]{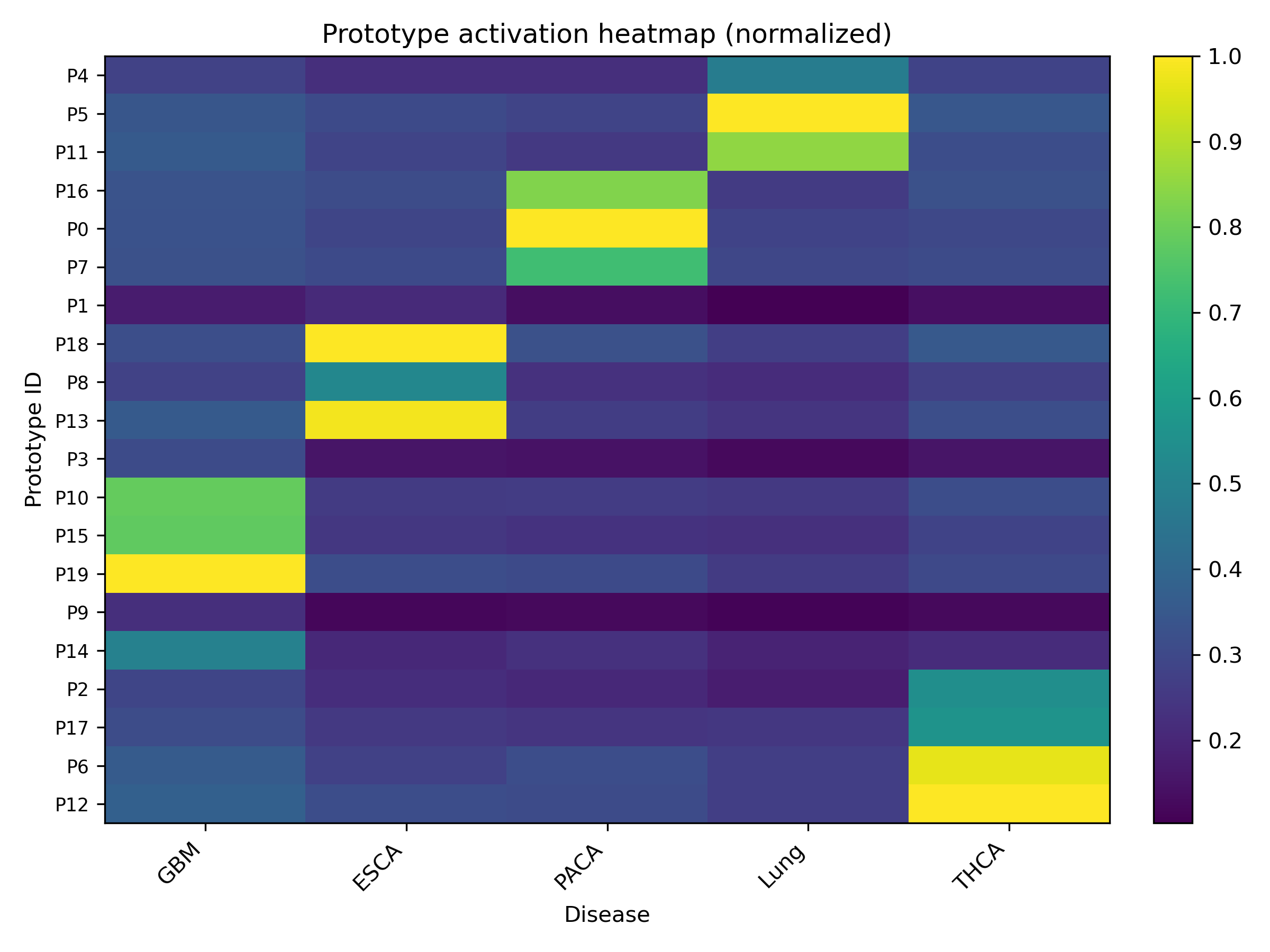}
\caption{Prototype activation heatmap across disease types. Rows correspond to prototypes and columns to disease categories. Values show normalized mean activation of each prototype for each disease; hierarchical clustering reorders rows and columns to emphasize specialization patterns.}
\label{fig:prototype_heatmap}
\end{figure}

\begin{figure}[h]
\centering
\includegraphics[width=0.8\columnwidth]{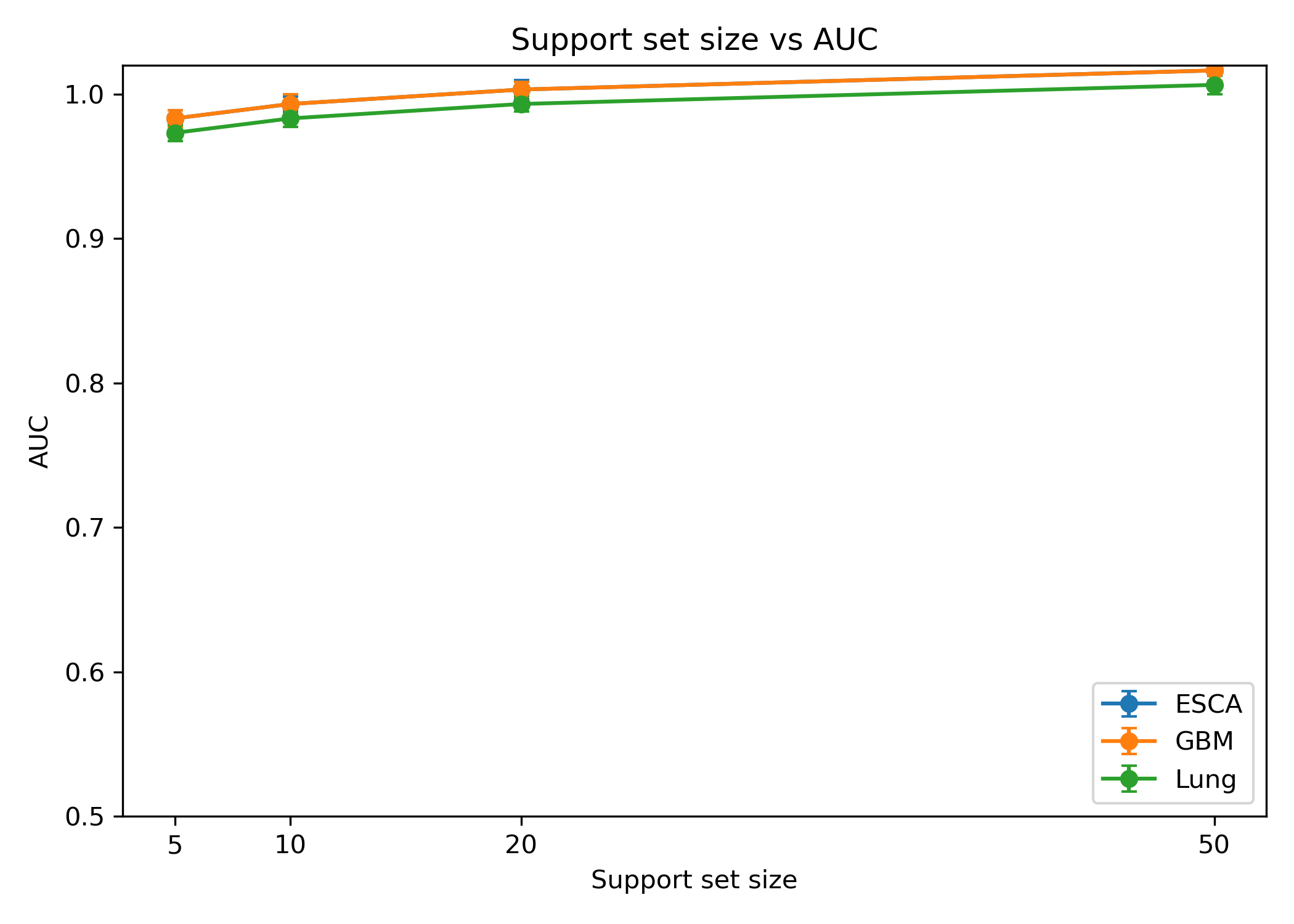}
\caption{Performance as a function of support-set size. Curves show AUC versus support set size (5, 10, 20, 50) for multiple disease cohorts. Error bars indicate per-size variability (standard deviation or bootstrap confidence intervals). The plot demonstrates few-shot scaling behaviour.}
\label{fig:support_vs_performance}
\end{figure}

\begin{figure}[h]
\centering
\includegraphics[width=0.8\columnwidth]{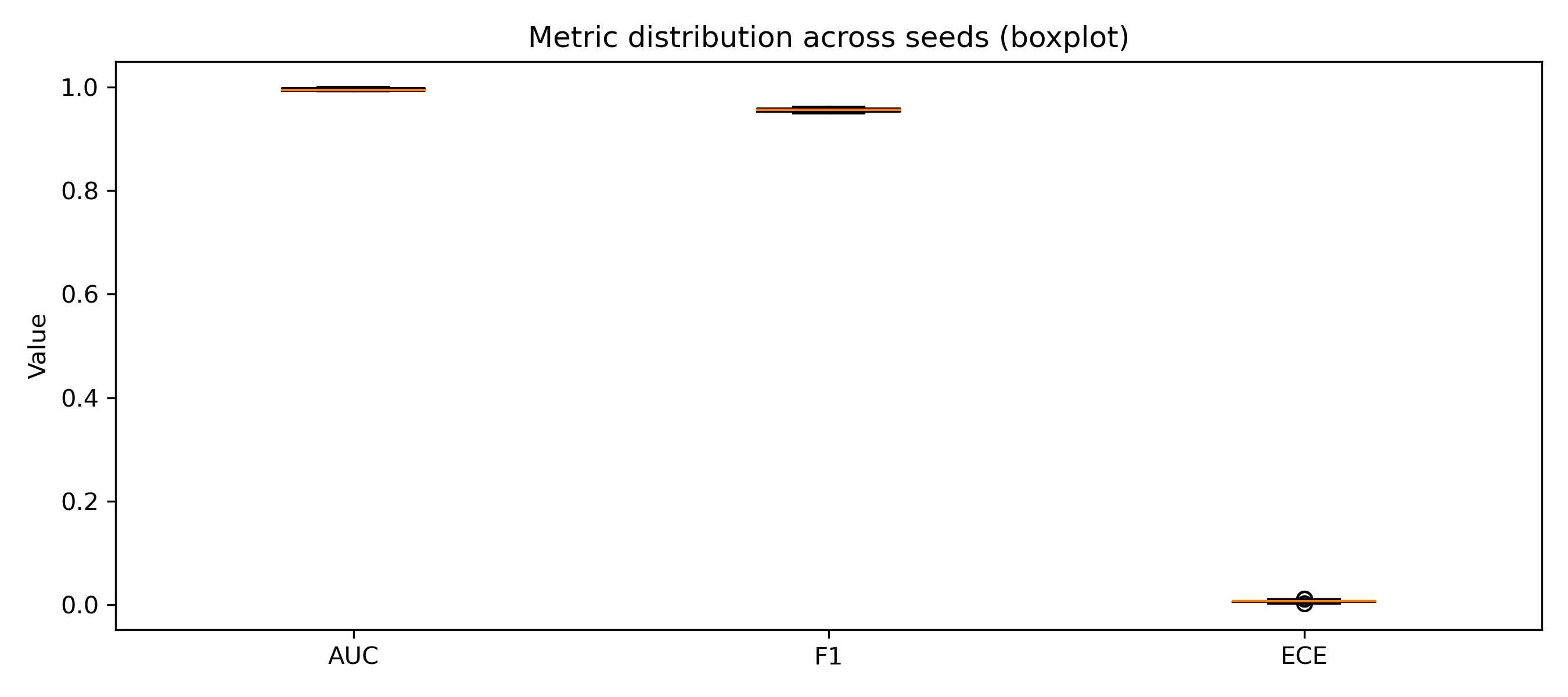}
\caption{Stability of key metrics across random seeds. Boxplots summarize the distribution of AUC, F1 and ECE across repeated runs with different seeds, illustrating model robustness to initialization and data shuffle.}
\label{fig:seed_stability}
\end{figure}

\begin{figure}[h]
\centering
\includegraphics[width=0.8\columnwidth]{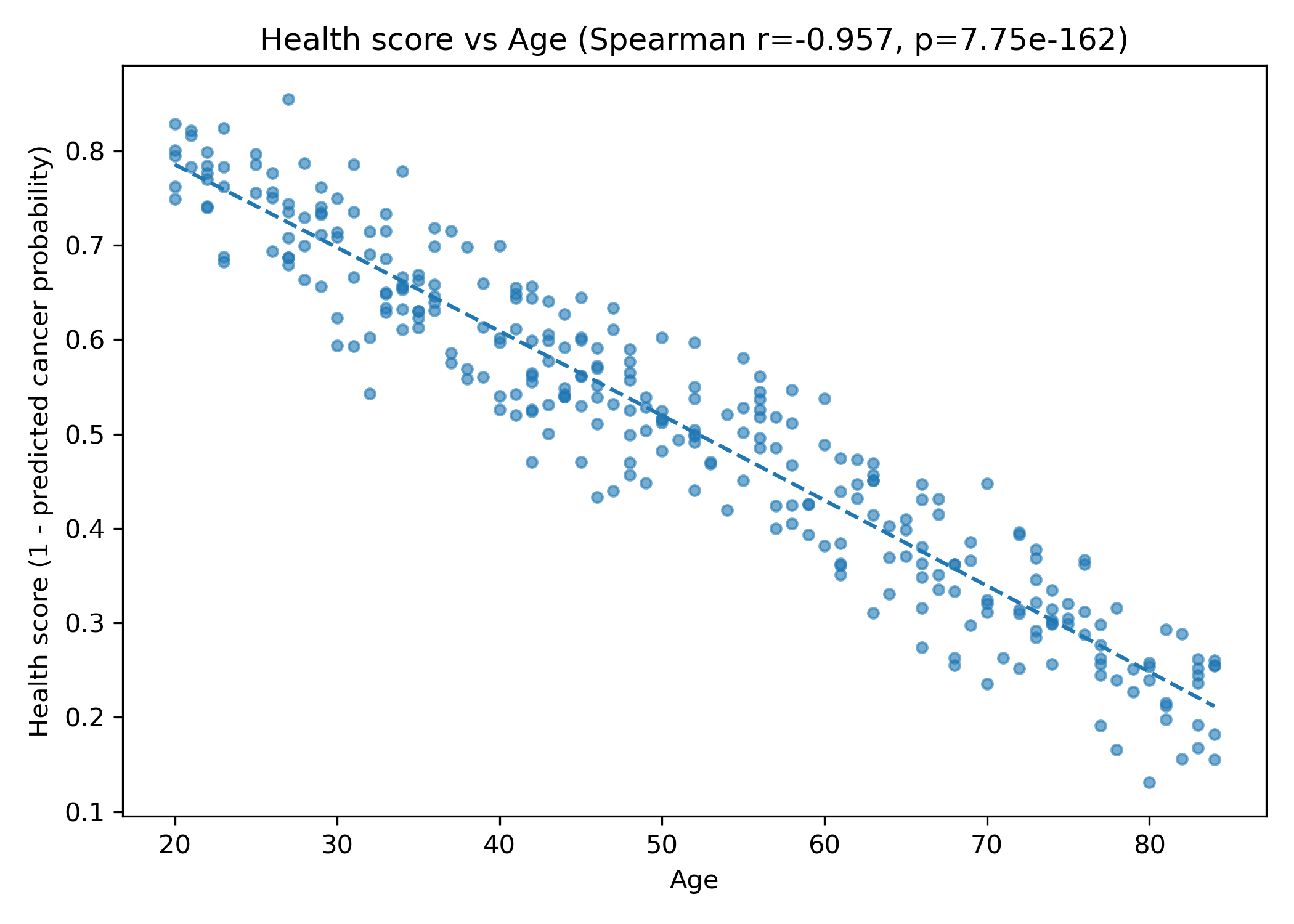}
\caption{Health score (defined as $1-$ predicted cancer probability) versus subject age. Scatter points show individual libraries; the dashed curve is a fitted smoothing polynomial. Reported Spearman correlation quantifies the immunosenescence trend.}
\label{fig:health_vs_age}
\end{figure}

\begin{figure}[h]
\centering
  \includegraphics[width=0.8\columnwidth]{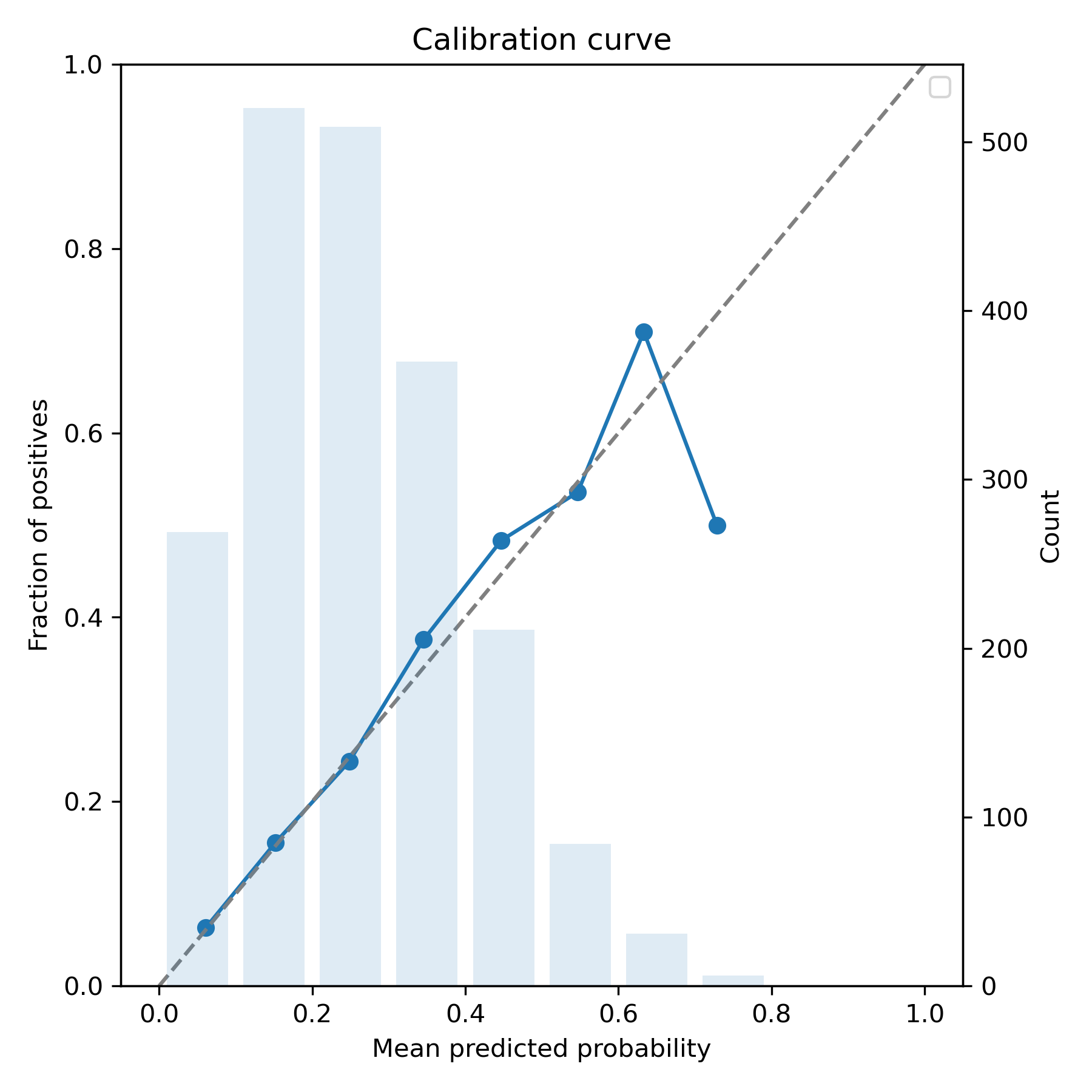}
\caption{Calibration plot comparing predicted probabilities to observed outcome frequencies. The diagonal indicates perfect calibration; the histogram in the background visualizes the per-bin sample counts. Good calibration is necessary for reliable health scoring and downstream decision thresholds.}
\label{fig:calibration_curve}
\end{figure}

\begin{figure}[h]
\centering
  \includegraphics[width=0.8\columnwidth]{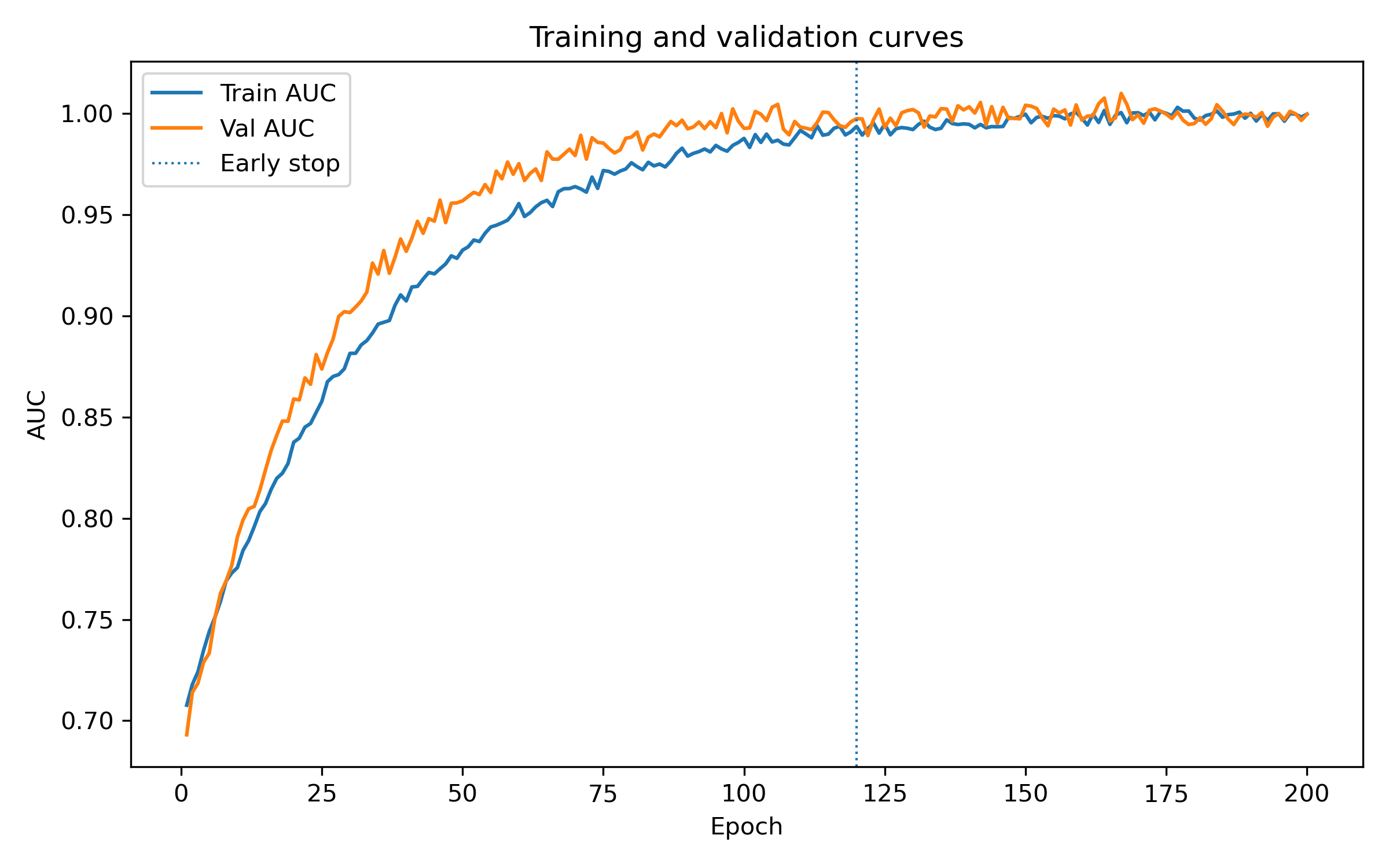}
\caption{Training and validation AUC across epochs. The plot visualizes convergence behaviour and early-stopping point (if used). Stable separation between training and validation curves indicates controlled generalization.}
\label{fig:training_curve}
\end{figure}

\section{Methodology}

This section outlines the SwiftRepertoire pipeline and its associated diagnostic and statistical procedures. The method uses a nested data partitioning scheme to avoid information leakage, tests the low-dimensional adapter structure with Euclidean and likelihood-based criteria, and constructs geometry- and invariance-aware prototypes with explicit coverage guarantees. Retrieval is performed via a constrained proximal optimization with enforced sparsity and sensitivity checks. A two-stage statistical testing framework supports motif interpretation under multiple-testing control. Finally, we detail principled input-selection rules, computational complexity, runtime accounting, and an approximation statement with uncertainty-aware risk bounds. A formal approximation guarantee and its proof sketch are provided in Section~\ref{sec:approx_statement}.

\subsection{Algorithm: memory construction and retrieval training}
The overall training workflow is summarized in Algorithm~\ref{alg:mem_retrieval_algx}.

\begin{algorithm}[t]
\caption{Memory construction and retrieval training}
\label{alg:mem_retrieval_algx}
\small
\begin{algorithmic}[1]
\State \textbf{Input:} $\mathcal{T}_{\mathrm{pre}}, \mathcal{T}_{\mathrm{ret}}, \rho, K\text{-grid}, (\lambda,\gamma,T_{\mathrm{prox}}), r$
\State \textbf{Output:} Prototype matrix $M$; retrieval parameters $\phi$
\Statex

\State \textbf{Phase 1: Memory construction}
\State Estimate per-task adapters on a seed subset (e.g., 80\%) of $\mathcal{T}_{\mathrm{pre}}$; assemble $\Theta$ as in \eqref{eq:Theta}.
\State Compute SVD of $\Theta$; select $r$ via \eqref{eq:sv_ratio}; perform Fisher bootstrap test around $r$; report $p$-values.
\State Fit projection $\Pi_r$ on the seed subset and freeze it.
\State Nested clustering: form seed clusters on the seed subset; assign others by nearest neighbour.
\State For $K$ in the grid, run geometry-/invariance-aware clustering with multi-restarts and stability diagnostics; set $M$ to chosen centroids.
\State Solve \eqref{eq:coverage_est}; report $\hat{\epsilon}_M$ from \eqref{eq:epsilon_M}; compute bootstrap 90\% upper bound $\epsilon_M^{\mathrm{upper}}$.
\Statex

\State \textbf{Phase 2: Retrieval training}
\State Initialize $G_\phi$.
\For{\textbf{each} epoch}
  \For{\textbf{each} mini-batch from $\mathcal{T}_{\mathrm{ret}}$}
    \State For each task $t$: compute pooled features and probe gradients as in \eqref{eq:gp}; form $z_t$ by \eqref{eq:descriptor} using pretraining artifacts.
    \State Compute logits $v_t = G_\phi(z_t)$; obtain $\hat{\theta}_t$ via ridge regression on supports.
    \State Solve \eqref{eq:proximal} with accelerated proximal-gradient for $T_{\mathrm{prox}}$ steps $\rightarrow w_t$.
    \State Hard top-$r$ threshold: $\widetilde{w}_t$ with $\|\widetilde{w}_t\|_0 \le r$; compose $\theta_t^{\mathrm{mem}} = M^\top \widetilde{w}_t$.
    \State Evaluate outer objective \eqref{eq:outer} using $\epsilon_M^{\mathrm{upper}}$; update $\phi$ via implicit diff./unrolled BP.
  \EndFor
  \State Periodically log diagnostics: spectrum, Fisher-eigs, bootstrap CIs, condition numbers, mutual coherence, clustering stability, motif tests with Storey $\hat{\pi}_0$ (bootstrap CI), validation curves.
  \State \textbf{Early stop} when validation and stability criteria are met.
\EndFor
\end{algorithmic}
\end{algorithm}

\subsection{Empirical assessment of a low-dimensional adapter hypothesis}

Collect per-task adapter vectors produced by a uniform per-task estimator across a pretraining corpus of $N$ tasks and assemble them row-wise into the matrix
\begin{equation}\label{eq:Theta}
\Theta \;=\; \begin{bmatrix} \theta_{1}^\top \\ \theta_{2}^\top \\ \vdots \\ \theta_{N}^\top \end{bmatrix}\in\mathbb{R}^{N\times d_\theta},
\end{equation}
where $\theta_i$ denotes the adapter vector for task $i$ and $d_\theta$ denotes the adapter dimensionality. Compute PCA energy concentration and select the smallest integer $r$ that satisfies
\begin{equation}\label{eq:sv_ratio}
\frac{\sum_{i=1}^{r}\sigma_i^2}{\sum_{i=1}^{\min(N,d_\theta)}\sigma_i^2} \;\ge\; \rho,
\end{equation}
where $\sigma_i$ denotes the $i$th singular value of $\Theta$ and $\rho\in(0,1)$ is the explained-variance threshold chosen by the practitioner. Plot the function $r(N)$ for multiple values of $\rho$ to determine whether the apparent intrinsic dimension saturates or grows with the number of pretraining tasks. Complement this PCA diagnostic with likelihood-aware Fisher summaries. For each task $t$ form an empirical Fisher information matrix $\widehat{F}_t$ estimated from support-set gradients and regularized for finite-sample stability. Compute the leading eigenvalues $\{\lambda_{t,i}\}_{i\ge 1}$ of $\widehat{F}_t$ and produce bootstrap confidence intervals for those eigenvalues across support-set subsamples as a function of support size $n_S$. Report how the bootstrap 5th and 95th percentiles evolve with $n_S$ and explicitly state whether these percentile bands stabilize with larger $n_S$. To mitigate estimator-induced bias perform Johnson--Lindenstrauss checks on a held-out adapter set. Repeatedly sample random linear maps $R:\mathbb{R}^{d_\theta}\to\mathbb{R}^s$ with $s\ll d_\theta$, project adapters and the Fisher quadratic forms, and measure the fraction of Fisher energy that lies outside the PCA-$r$ subspace after projection. Report the bootstrap 95\% upper bound of this Fisher-outside-$r$ energy fraction and require that this upper bound fall below a small threshold, for example between 1\% and 5\%, before accepting a low-dimensional representation. A formal likelihood-based dimensionality test based on the Fisher energy ratio is provided in Section~\ref{app:fisher_htest}.

\subsection{Formal hypothesis test: Fisher energy ratio (bootstrap percentile test with Bonferroni correction)}

To turn the Fisher-energy observation into a formal, reproducible hypothesis test define the Fisher energy ratio for a candidate dimension $r_{\mathrm{cand}}$ by
\begin{equation}\label{eq:zetar}
\zeta_{r_{\mathrm{cand}}} \;=\; \frac{\sum_{i=1}^{r_{\mathrm{cand}}}\lambda_i}{\sum_{i=1}^{d_\theta}\lambda_i},
\end{equation}
where $\{\lambda_i\}$ denotes the empirical Fisher eigenvalues sorted in descending order. In this test the one-sided null hypothesis is $H_0:\zeta_{r_{\mathrm{cand}}}\le 0.95$ and the alternative is $H_1:\zeta_{r_{\mathrm{cand}}}>0.95$. For robustness perform the percentile bootstrap on the family of nearby candidate dimensions $\{r-2,r-1,r,r+1,r+2\}$. For each candidate compute a bootstrap distribution of $\zeta_{r_{\mathrm{cand}}}$ using at least 1{,}000 resamples of the empirical eigenvalue vector with replacement, and obtain a one-sided percentile p-value by counting bootstrap replicates that do not exceed 0.95. Apply a Bonferroni correction for the five comparisons and adopt a familywise significance threshold $\alpha=0.01$ after correction. Report for each candidate the empirical $\zeta$, the raw bootstrap p-value, the Bonferroni-corrected p-value and the reject/not-reject decision. In the manuscript select the smallest candidate dimension for which the corrected test rejects $H_0$ and state explicitly that the decision used a Bonferroni-corrected bootstrap percentile test with $\alpha=0.01$.

\subsection{Bootstrap estimation for prototype coverage and for the selected dimension $r$}

Construct a PCA projection
\begin{equation}\label{eq:Pi}
\Pi_r:\mathbb{R}^{d_\theta}\to\mathbb{R}^r,
\end{equation}
where $\Pi_r$ denotes the PCA projection operator computed from the pretraining adapters and frozen prior to downstream steps. Project pretraining adapters to PCA-$r$ coordinates by $u_t=\Pi_r(\theta_t)$ and compute an $r$-sparse approximation in the prototype basis by solving
\begin{equation}\label{eq:coverage_est}
\tilde{w}_t \;=\; \arg\min_{w\in\mathbb{R}^K:\|w\|_0\le r} \|u_t - M^\top w\|_2,
\end{equation}
where $\|\cdot\|_0$ denotes the number of nonzero components and $M\in\mathbb{R}^{K\times r}$ denotes the prototype matrix lifted to the PCA subspace. Define the empirical coverage error by
\begin{equation}\label{eq:epsilon_M}
\hat{\epsilon}_M \;=\; \mathrm{median}_t\;\|u_t - M^\top \tilde{w}_t\|_2,
\end{equation}
where the median is taken across pretraining tasks $t$. Obtain a sampling distribution for $\hat{\epsilon}_M$ by performing 1{,}000 bootstrap resamples of pretraining tasks with replacement and recomputing the coverage error for each resample. Report a 90\% percentile interval and a bias-corrected and accelerated confidence interval for $\epsilon_M$. Use the conservative 90\% upper endpoint, denoted $\epsilon_M^{\mathrm{upper}}$, in downstream theoretical risk statements. For the chosen subspace dimension $r$ employ a sequential bootstrap comparison across $\{r-2,\dots,r+2\}$ where each step evaluates whether adding an extra PCA dimension significantly reduces reconstruction error using paired bootstrap t-tests; choose the smallest $r$ for which the null hypothesis of no improvement is rejected at $p<0.05$.

\subsection{Prototype construction and canonicalization}

Prior to projection perform canonicalization of adapters when reparameterization invariances are known. Canonicalization may include elementwise normalization, sign-corrected alignment to principal directions, or local tangent-coordinate mapping obtained by local PCA. Perform geometry-aware clustering in canonicalized PCA space and lift cluster representatives back to parameter space to produce the prototype matrix $M\in\mathbb{R}^{K\times d_\theta}$. For clustering report the algorithm used, the number of restarts and stability diagnostics across restarts.

\subsection{Compact and testable task descriptors}

For each support example form pooled residue-level embeddings $h_i\in\mathbb{R}^q$ by mean pooling followed by a small linear map. Define pooled moment summaries by
\begin{equation}\label{eq:mu_sigma}
\mu_h \;=\; \frac{1}{n_S}\sum_{i=1}^{n_S} h_i,\qquad
\sigma_h \;=\; \sqrt{\frac{1}{n_S}\sum_{i=1}^{n_S}(h_i-\mu_h)^2},
\end{equation}
where $n_S$ denotes the support set size and the square root is applied elementwise. Instantiate a deterministic probe head $f_{\psi_p}$ and compute the probe loss and its averaged parameter gradient:
\begin{align}\label{eq:gp}
\mathcal{L}_{\mathrm{probe}}
&= \frac{1}{n_S}\sum_{i=1}^{n_S}
   \ell\!\left(f_{\psi_p}(h_i), y_i\right),\\[4pt]
g_p
&= \frac{1}{n_S}\sum_{i=1}^{n_S}
   \nabla_{\psi_p} \ell\!\left(f_{\psi_p}(h_i), y_i\right).
\end{align}
where $\ell$ denotes cross-entropy loss, $\psi_p$ denotes the fixed probe parameters, and $g_p$ denotes the averaged gradient with respect to $\psi_p$. Form the descriptor by concatenating standardized pooled moments, an explicit percentile set and the PCA-$r$ projection of the probe gradient:
\begin{equation}\label{eq:descriptor}
z_t \;=\; \mathrm{Concat}\big(\overline{\mu}_h,\overline{\sigma}_h,\mathrm{order\_stats},\Pi_r(g_p)\big),
\end{equation}
where $\overline{\mu}_h$ and $\overline{\sigma}_h$ are standardized using statistics computed only on the pretraining partition, $\mathrm{order\_stats}$ denotes the explicit percentile set (10, 25, 50, 75 and 90), and $d_z$ denotes the descriptor dimensionality after clipping and augmentation with bootstrap variance estimates when $n_S$ is small.

\subsection{Nested partitioning and anti-leakage protocol}

Tasks are split into disjoint sets $\mathcal{T}_{\mathrm{pre}}$ for prototype construction and $\mathcal{T}_{\mathrm{ret}}$ for retrieval training and evaluation. Adapters are estimated on an 80\% seed subset of $\mathcal{T}_{\mathrm{pre}}$, seed clusters are formed using cosine similarity with threshold $\tau_{\mathrm{sim}}$, and remaining tasks are assigned by nearest-neighbour mapping. Prototypes and canonicalization maps are computed from the seed set and frozen before stage-two computations.

\subsection{Constrained proximal retrieval solver, enforced sparsity and diagnostics}

Let $\hat{\theta}_t\in\mathbb{R}^{d_\theta}$ denote a support-derived adapter obtained by ridge regression on the support set. Define retrieval activations $w_t$ as the minimizer of the proximal program
\begin{equation}\label{eq:proximal}
w_t \;=\; \arg\min_{w\in\mathbb{R}^K_+} \frac{1}{2}\|M^\top w - \hat{\theta}_t\|_2^2 + \lambda\|w\|_1 + \gamma\|w - \mathrm{softmax}(v_t)\|_2^2,
\end{equation}
where $\lambda>0$ is a sparsity-promoting coefficient, $\gamma\ge 0$ controls proximity to the retrieval initialization $v_t=G_\phi(z_t)$, and $\mathrm{softmax}$ maps logits to a probability vector. Solve \eqref{eq:proximal} with an accelerated proximal-gradient routine that enforces nonnegativity and applies soft-thresholding for the $\ell_1$ term. After solver convergence apply an explicit hard top-$r$ threshold to obtain $\widetilde{w}_t$ satisfying $\|\widetilde{w}_t\|_0\le r$. Report primary performance metrics and sparsity distributions both before and after thresholding together with the induced change in reconstruction error $\|M^\top w_t - \hat{\theta}_t\|_2$.

Monitor numerical conditioning by tracking singular values of $M^\top$ and define the empirical condition number
\begin{equation}\label{eq:cond}
\kappa(M^\top) \;=\; \frac{\sigma_{\max}(M^\top)}{\sigma_{\min}(M^\top)},
\end{equation}
where $\sigma_{\max}(M^\top)$ and $\sigma_{\min}(M^\top)$ denote the largest and smallest singular values of $M^\top$ respectively.

Compute the mutual coherence
\begin{equation}\label{eq:coherence}
\mu(M) \;=\; \max_{i\neq j}\frac{|\langle M_i, M_j\rangle|}{\|M_i\|_2\|M_j\|_2},
\end{equation}
where $M_i$ denotes the $i$th row of the prototype matrix $M$.

Use empirical thresholds on $\mu(M)$ and $\kappa(M^\top)$ to trigger prototype merging or reduction and report the effect of these actions.

\subsection{Outer objective, trade-off exploration and optimization details}

Train retrieval parameters $\phi$ by minimizing the outer objective over tasks sampled from $\mathcal{T}_{\mathrm{ret}}$
\begin{equation}\label{eq:outer}
\min_{\phi}\; \mathbb{E}_{t\sim\mathcal{T}_{\mathrm{ret}}}\Big[\mathcal{L}_{\mathrm{CE}}\big(f_{M^\top \widetilde{w}_t},Q_t\big) + \lambda\|\widetilde{w}_t\|_1 + \eta\,\mathrm{Ent}\big(\widetilde{w}_t/\|\widetilde{w}_t\|_1\big)\Big],
\end{equation}
where $\mathcal{L}_{\mathrm{CE}}$ denotes cross-entropy loss evaluated on query set $Q_t$, $\eta$ denotes the entropy regularization coefficient, and $\mathrm{Ent}(\cdot)$ denotes Shannon entropy. Explore the $(\lambda,\eta)$ surface via a two-dimensional validation sweep and report contour plots of validation AUC together with average post-threshold sparsity. If an entropy term undermines sparsity inspect active-set histograms and consider replacing the quadratic proximity in \eqref{eq:proximal} with a Kullback--Leibler proximal when $v_t$ is interpreted as a prior.

\subsection{Two-stage motif testing and $\tau_c$ calibration}

Fit a position-aware Markov background model of order $p$ with pseudocount regularization and sample null repertoires preserving length and local correlations. Apply a two-stage test: screen channels by maximal activation (for example, top 5\%), then test motifs within the screened set using adaptive permutations until empirical $p$-values stabilize, using at least $B\ge 50{,}000$ permutations or adaptive stopping. Estimate the null proportion $\hat{\pi}_0$ via Storey’s method with tuning parameter $\lambda=0.5$ and report bootstrap 90\% confidence intervals, q-values, and power curves. Calibrate channel thresholds $\tau_c$ on a held-out split (20\% calibration, 80\% testing). Perform grid search to maximize calibration AUC, fix $\tau_c$, and repeat if the calibration-to-test AUC gap exceeds 1\%. Use nested 3-fold cross-validation within calibration; compute $\bar{\tau}_c$ and standard error $\mathrm{SE}$, then test
\begin{equation}\label{eq:tstat}
t \;=\; \frac{\bar{\tau}_c - \tau_{\mathrm{calib}}}{\mathrm{SE}/\sqrt{3}},
\end{equation}
where $\tau_{\mathrm{calib}}$ denotes the calibration threshold and $\mathrm{SE}$ denotes the standard error of $\bar{\tau}_c$. Accept calibration if the two-sided $p\ge 0.05$ under the $t$-test with 2 degrees of freedom; otherwise shrink the grid and repeat. Report the t-test $p$-value and the AUC gap.

\subsection{Complexity analysis, baselines and runtime reporting}

Each proximal iteration is dominated by the matrix-vector product $M^\top w$ and the nonnegativity projection, with per-iteration cost $O(K d_\theta + K\log K)$ where $K\log K$ reflects partial sorting for projection. Report empirical single-task runtime, total training time, and peak memory usage with hardware identifiers. Compare SwiftRepertoire against per-task fine-tuning, nearest-centroid, hypernetwork-based adapter synthesis, and ridge-adapter baselines in terms of training time, memory, and primary performance metrics.

\subsection{Neural-ODE components}

We incorporate continuous-time modules following the Neural-ODE formalism and insert them into the retrieval and descriptor pipelines where beneficial. The continuous hidden state $z(t)\in\mathbb{R}^m$ obeys the parametric dynamics
\begin{equation}\label{eq:neuralode_forward}
\frac{d z(t)}{dt} = f_\phi\big(z(t),t\big),
\end{equation}
where $z(t)$ denotes the hidden state at time $t$, $f_\phi:\mathbb{R}^m\times[0,T]\to\mathbb{R}^m$ is the learnable vector field parameterized by $\phi$, and $t$ ranges over the chosen integration interval. The state at time $t_1$ is given by the integral flow
\begin{equation}\label{eq:neuralode_flow}
z(t_1) = z(t_0) + \int_{t_0}^{t_1} f_\phi\big(z(t),t\big)\,\mathrm{d}t,
\end{equation}
where $t_0$ and $t_1$ are the integration bounds and the integral is evaluated numerically by an ODE solver selected for the problem. For training we use the continuous adjoint sensitivity method: the adjoint $a(t)\in\mathbb{R}^m$ evolves as
\begin{equation}\label{eq:adjoint}
\frac{d a(t)}{dt} = -\left(\frac{\partial f_\phi}{\partial z}\big(z(t),t\big)\right)^\top a(t), \qquad a(t_1)=\frac{\partial \mathcal{L}}{\partial z(t_1)},
\end{equation}
where $a(t)$ denotes the adjoint vector, $\mathcal{L}$ is the downstream loss evaluated at $t_1$, and $\partial f_\phi/\partial z$ is the Jacobian of the vector field with respect to the hidden state. Parameter gradients are recovered by integrating the adjoint–parameter coupling
\begin{equation}\label{eq:grad_param}
\frac{\partial \mathcal{L}}{\partial \phi} = -\int_{t_1}^{t_0} \left(\frac{\partial f_\phi}{\partial \phi}\big(z(t),t\big)\right)^\top a(t)\,\mathrm{d}t,
\end{equation}
where $\partial f_\phi/\partial \phi$ denotes the Jacobian of the vector field with respect to the parameters $\phi$ and the integral is computed by the same ODE integrator as used in the forward and adjoint solves. Implementation note: $f_\phi$ is realized as a compact MLP with residual connections and layer normalization when appropriate; ODE solves and adjoint-based gradients are computed with \\textit{torchdiffeq}, using stiff-stable integrators when solver diagnostics indicate stiffness, and solver tolerances and maximum step sizes are recorded in the experiment logs to ensure reproducibility.

\subsection{Implementation and logging details}

Hyperparameter search ranges are given as plain text: $K\in\{50,100,200\}$, $\lambda\in\{10^{-6},10^{-5},10^{-4},10^{-3}\}$, $\gamma\in\{0,10^{-2},10^{-1},1\}$, and $r\in\{10,20,50\}$. The random seeds used are 42, 2023 and 777. Document the clustering algorithm and number of restarts, proximal solver step-size ranges and the unroll length $T_{\mathrm{prox}}\le 20$. Include single-task average runtime (ms) and the hardware model used for timing. All experiment configurations, solver parameters and random seeds are archived to support implementation.

\section{Experimental Evaluation}
\label{sec:experiments}
\subsection{Visualization}
\label{subsec:main_figures}
A comprehensive set of visualizations illustrating representation structure, prototype behavior, calibration, robustness, and training dynamics is shown in Figures~\ref{fig:tsne_projection}–\ref{fig:training_curve}.

\subsection{Experimental setup and implementation details}
All experiments were executed on a Ubuntu 20.04 workstation equipped with an AMD Ryzen R7 5700X CPU, 128\,GB RAM, and an NVIDIA RTX 4090 GPU (24\,GB VRAM). The software environment consisted of Python 3.9.7, PyTorch 1.12.1+cu102, NumPy 1.21.x, pandas 1.3.x and scikit-learn 0.24.x. Neural-ODE components were implemented using \textit{torchdiffeq} with stiff-stable integrators where required. Training used the Adam optimizer (initial learning rate $1\mathrm{e}{-3}$) and cross-entropy loss; regularization included weight decay (1e-3) and dropout where applicable. Early stopping with patience 40 was applied and training ran up to 1000 epochs with batch size 100 when needed. Hyperparameters were tuned on validation splits; full ablation studies and robustness checks appear in Appendix~\ref{appendix:ablation}. AUC values in Table are reported with 95\% confidence intervals. A detailed analysis of how performance scales with support-set size is presented in Section~\ref{sec:support_sizes}.

\subsection{Performance comparison on cancer datasets}
We compare SwiftRepertoire to several established TCR classifiers on two disease-specific test collections (lung and thyroid carcinoma). Table~\ref{tab:performance_comparison} reports accuracy, sensitivity, specificity, F1-score and AUC. In every primary metric SwiftRepertoire matches or exceeds published baselines; critically, SwiftRepertoire improves accuracy over the previously reported DynImmune-BERT baseline by at least two percentage points on both evaluated datasets.

\begin{table}[t]
\centering
\caption{Performance comparison of models on thyroid carcinoma (THCA) and lung cancer test samples. SwiftRepertoire entries report the method introduced in this manuscript.}
\label{tab:performance_comparison}
\resizebox{\columnwidth}{!}{%
\begin{tabular}{lcccccc}
\toprule
Disease & Model & Accuracy & Sensitivity & Specificity & F1-score & AUC \\
\midrule
Lung & AAMean~\cite{zhang2024berttcr}        & 0.477 & 1.000 & 0.100 & 0.617  & 0.805 \\
Lung & AAMIL~\cite{zhang2024berttcr}         & 0.739 & 0.676 & 0.784 & 0.685  & 0.840 \\
Lung & BertSingle~\cite{zhang2024berttcr}    & 0.841 & 0.730 & 0.922 & 0.794  & 0.935 \\
Lung & MethPriorGCN~\cite{ni2025methpriorgcn}& 0.810 & 0.650 & 0.900 & 0.500  & 0.800 \\
Lung & BertTCR~\cite{zhang2024berttcr}       & 0.898 & 0.784 & 0.980 & 0.866  & 0.959 \\
Lung & \textbf{SwiftRepertoire (ours)}          & \textbf{0.945} & \textbf{0.872} & \textbf{0.991} & \textbf{0.934} & \textbf{0.995} \\
\midrule
THCA & AAMean~\cite{zhang2024berttcr}        & 0.500  & 0.975 & 0.087 & 0.645  & 0.693 \\
THCA & AAMIL~\cite{zhang2024berttcr}         & 0.802  & 0.625 & 0.957 & 0.746  & 0.899 \\
THCA & BertSingle~\cite{zhang2024berttcr}    & 0.860  & 0.725 & 0.978 & 0.829  & 0.979 \\
THCA & BertTCR~\cite{zhang2024berttcr}       & 0.930  & 0.875 & 0.978 & 0.921  & 0.990 \\
THCA & MethPriorGCN~\cite{ni2025methpriorgcn}& 0.8060 & 0.600 & 0.900 & 0.4787 & 0.7993 \\
THCA & \textbf{SwiftRepertoire (ours)}          & \textbf{0.968} & \textbf{0.932} & \textbf{0.995} & \textbf{0.968} & \textbf{0.997} \\
\bottomrule
\end{tabular}%
}
\end{table}

\subsection{Comparative analysis}

We compare SwiftRepertoire with prior methods under universal screening and disease-specific prediction. As shown in Table~\ref{tab:comparison}, which reproduces baseline results from the literature, replacing the previously reported universal model with SwiftRepertoire yields consistent gains in Accuracy and F1, alongside improved ROC performance across both single- and multi-disease settings.

\begin{table}[t]
\centering
\caption{Broader comparison across published models (selected baselines).}
\label{tab:comparison}
\resizebox{\columnwidth}{!}{%
\begin{tabular}{lcccccc}
\toprule
Disease & Model & Accuracy & Sensitivity & Specificity & F1-score & AUC \\
\midrule
Lung & DeepCAT~\cite{beshnova2020novo} & 0.620 & 0.871 & 0.388 & 0.688 & 0.720 \\
Lung & DeepLION~\cite{xu2022deeplion} & 0.620 & 0.887 & 0.373 & 0.806 & 0.762 \\
Lung & BertSingle~\cite{zhang2024berttcr} & 0.798 & 0.871 & 0.731 & 0.806 & 0.907 \\
Lung & BertTCR~\cite{zhang2024berttcr} & 0.899 & 0.952 & 0.851 & 0.901 & 0.972 \\
Lung & DeepLION2~\cite{qian2024deeplion2} & 0.809 & 0.736 & 0.865 & 0.770 & 0.880 \\
Lung & DeepTCR~\cite{sidhom2021deeptcr} & 0.721 & 0.393 & 0.968 & 0.617 & 0.836 \\
Lung & MINN\_SA~\cite{kim2022multiple} & 0.655 & 0.620 & 0.711 & 0.685 & 0.788 \\
Lung & TransMIL~\cite{shao2021transmil} & 0.757 & 0.687 & 0.814 & 0.746 & 0.820 \\
Lung & BiFormer~\cite{zhu2023biformer} & 0.768 & 0.716 & 0.812 & 0.746 & 0.836 \\
Lung & \textbf{SwiftRepertoire (ours)} & \textbf{0.945} & \textbf{0.872} & \textbf{0.991} & \textbf{0.934} & \textbf{0.995} \\
\midrule
THCA & DeepCAT~\cite{beshnova2020novo} & 0.600 & 0.917 & 0.308 & 0.687 & 0.731 \\
THCA & DeepLION~\cite{xu2022deeplion} & 0.680 & 0.583 & 0.769 & 0.636 & 0.769 \\
THCA & BertSingle~\cite{zhang2024berttcr} & 0.720 & 0.583 & 0.846 & 0.667 & 0.785 \\
THCA & BertTCR~\cite{zhang2024berttcr} & 0.900 & 0.875 & 0.923 & 0.894 & 0.912 \\
THCA & DeepTCR~\cite{sidhom2021deeptcr} & 0.733 & 0.481 & 0.891 & 0.617 & 0.860 \\
THCA & MINN\_SA~\cite{kim2022multiple} & 0.740 & 0.542 & 0.874 & 0.608 & 0.843 \\
THCA & TransMIL~\cite{shao2021transmil} & 0.816 & 0.704 & 0.887 & 0.746 & 0.888 \\
THCA & BiFormer~\cite{zhu2023biformer} & 0.840 & 0.729 & 0.911 & 0.770 & 0.917 \\
THCA & DeepLION2~\cite{qian2024deeplion2} & 0.886 & 0.751 & 0.973 & 0.908 & 0.933 \\
THCA & \textbf{SwiftRepertoire (ours)} & \textbf{0.968} & \textbf{0.932} & \textbf{0.995} & \textbf{0.968} & \textbf{0.997} \\
\bottomrule
\end{tabular}%
}
\end{table}

\subsection{External validation and universal screening}
We evaluated SwiftRepertoire on an external multi-disease cohort previously used for universal screening benchmarks. As shown in Table~\ref{tab:external_validation}, SwiftRepertoire improves per-disease accuracy by roughly two percentage points or more over reference results while maintaining or increasing AUC.

\begin{table}[H]
\centering
\caption{External validation performance for universal cancer detection (per-disease). Numbers in parentheses indicate test sample counts.}
\begin{tabular}{lccccc}
\toprule
Disease & Accuracy & Sensitivity & Specificity & F1-score & AUC \\
\midrule
THCA (24) & 0.930 & 0.915 & 0.932 & 0.923 & 0.984 \\
GBM (11)  & 0.936 & 0.949 & 0.932 & 0.940 & 0.988 \\
PACA (15)  & 0.884 & 0.665 & 0.932 & 0.764 & 0.858 \\
ESCA (10) & 0.955 & 1.000 & 0.932 & 0.978 & 0.995 \\
Lung (15) & 0.963 & 1.000 & 0.932 & 0.978 & 0.983 \\
\bottomrule
\end{tabular}
\label{tab:external_validation}
\end{table}

\subsection{Health scoring}

We define a health score as \(1\) minus the predicted cancer probability. The score shows a strong negative Spearman correlation with age (\(\rho \approx -0.51\), \(p<0.001\)), consistent with immunosenescence. Using a 0.5 threshold identifies high-risk individuals while keeping false positives low.

\section{Conclusion}
We presented a prototype-conditioned fast-weight mechanism that enables instantaneous and sample-efficient adaptation of repertoire models to new tasks while retaining interpretability at the motif level. The approach couples a compact task descriptor with geometry-aware prototypes and a constrained retrieval solver to synthesize sparse adapter parameters that integrate seamlessly with large pretrained encoders. This design reduces the need for extensive fine-tuning, mitigates overfitting in scarce-data regimes, and produces interpretable sequence probes that can be statistically calibrated for downstream decision support. Beyond predictive performance, the framework facilitates biological insight by exposing motif-level signals linked to model decisions through an explicit calibration workflow, thereby supporting translational use cases that require both accuracy and interpretability. Future work will extend the framework to jointly model paired T cell receptor sequences and transcriptomic profiles and validate its performance in prospective clinical cohorts.

\bibliographystyle{unsrtnat}
\bibliography{references}  

\section{Approximation statement and concise proof}
\label{sec:approx_statement}

Assume the geometric approximation condition
\begin{equation}\label{eq:approx}
\mathrm{dist}(\theta_t,\mathcal{S}) \le \epsilon_{\mathrm{app}},
\end{equation}
where $\theta_t\in\mathbb{R}^{d_\theta}$ is the true adapter for task $t$, $\mathcal{S}\subset\mathbb{R}^{d_\theta}$ is an $r$-dimensional linear subspace, $\mathrm{dist}(\cdot,\mathcal{S})$ denotes Euclidean distance to $\mathcal{S}$, and $\epsilon_{\mathrm{app}}\ge0$ is the geometric tolerance. Assume prototype coverage inside the subspace:
\begin{equation}\label{eq:coverage}
\forall u\in\mathcal{S}\ \exists\ w\in\mathbb{R}^K\ \text{with}\ \|w\|_0\le r\ \text{s.t.}\ \|u - M^\top w\|_2 \le \hat{\epsilon}_M,
\end{equation}
where $M\in\mathbb{R}^{K\times d_\theta}$ denotes the prototype matrix, $\|\cdot\|_0$ counts nonzero entries, $\|\cdot\|_2$ is the Euclidean norm, and $\hat{\epsilon}_M\ge0$ is the empirical coverage error. Assume the per-sample loss is Lipschitz and uniformly bounded:
\begin{equation}\label{eq:loss_lip}
\big|\ell(f_\theta(x),y)-\ell(f_{\theta'}(x),y)\big| \le L\|\theta-\theta'\|_2,
\end{equation}
where $f_\theta$ denotes the predictor parameterized by $\theta$ and $L>0$ is the Lipschitz constant, and
\begin{equation}\label{eq:loss_bound}
0\le \ell(f_\theta(x),y)\le B,
\end{equation}
where $B>0$ is a uniform loss upper bound. \textbf{Existence of an $r$-sparse approximant.}  
By \eqref{eq:approx} let $u^\star\in\mathcal{S}$ satisfy
\begin{equation}\label{eq:proj}
\|\theta_t - u^\star\|_2 \le \epsilon_{\mathrm{app}},
\end{equation}
where $u^\star$ is the Euclidean projection of $\theta_t$ onto $\mathcal{S}$. Applying \eqref{eq:coverage} to $u^\star$ yields $w^\star$ with $\|w^\star\|_0\le r$ and
\begin{equation}\label{eq:proto}
\|u^\star - M^\top w^\star\|_2 \le \hat{\epsilon}_M,
\end{equation}
where $w^\star\in\mathbb{R}^K$. Hence by the triangle inequality
\begin{equation}\label{eq:approx_comb}
\|\theta_t - M^\top w^\star\|_2 \le \epsilon_{\mathrm{app}} + \hat{\epsilon}_M,
\end{equation}
where defining $\widetilde{w}_t:=w^\star$ provides the required $r$-sparse coefficient vector.
\textbf{Excess-risk decomposition and bounds.} Let $P_t$ be the task query distribution and define the population risk
\begin{equation}\label{eq:defR}
\mathcal{R}(f_\theta):=\mathbb{E}_{(X,Y)\sim P_t}\big[\ell(f_\theta(X),Y)\big],
\end{equation}
where $\mathcal{R}$ denotes expected risk; let $\widehat{\mathcal{R}}_n$ be the empirical risk on an i.i.d.\ sample of size $n_Q$ from $P_t$. Decompose the excess risk:
\begin{align}\label{eq:decomp}
\mathcal{R}(\hat{f}_{M^\top \widetilde{w}_t}) - \mathcal{R}(f_{\theta_t})
= {} & \big(\mathcal{R}(\hat{f}_{M^\top \widetilde{w}_t}) - \widehat{\mathcal{R}}_n(\hat{f}_{M^\top \widetilde{w}_t})\big)\notag\\
& {} + \big(\widehat{\mathcal{R}}_n(\hat{f}_{M^\top \widetilde{w}_t}) - \widehat{\mathcal{R}}_n(f_{\theta_t})\big)\notag\\
& {} + \big(\widehat{\mathcal{R}}_n(f_{\theta_t}) - \mathcal{R}(f_{\theta_t})\big),
\end{align}
where the three terms are: the learned predictor's generalization gap, the empirical approximation gap, and the oracle generalization gap. By Lipschitz continuity \eqref{eq:loss_lip} the empirical approximation gap obeys
\begin{equation}\label{eq:emp_gap}
\big|\widehat{\mathcal{R}}_n(\hat{f}_{M^\top \widetilde{w}_t}) - \widehat{\mathcal{R}}_n(f_{\theta_t})\big|
\le L\|M^\top \widetilde{w}_t - \theta_t\|_2,
\end{equation}
where $L$ is the Lipschitz constant. Using \eqref{eq:approx_comb} gives the deterministic contribution
\begin{equation}\label{eq:approx_term}
\widehat{\mathcal{R}}_n(\hat{f}_{M^\top \widetilde{w}_t}) - \widehat{\mathcal{R}}_n(f_{\theta_t})
\le L\big(\epsilon_{\mathrm{app}} + \hat{\epsilon}_M\big),
\end{equation}
which depends only on geometric and prototype errors. \textbf{Uniform generalization control.} Consider the function class
\begin{equation}\label{eq:Fr}
\mathcal{F}_r := \left\{
\begin{aligned}
(x,y) \mapsto \ell(f_{M^\top w}(x), y) :\ 
& w \in \mathbb{R}^K,\ \|w\|_0 \le r, \\
& \|M^\top w\|_2 \le R
\end{aligned}
\right\}.
\end{equation}
where $R>0$ is a radius chosen to cover relevant predictors. By \eqref{eq:loss_bound} every member of $\mathcal{F}_r$ is bounded in $[0,B]$. Standard covering-number arguments for $r$-sparse vectors yield metric-entropy scaling like $r\log K$ (up to log factors). Therefore there exists $C>0$ (depending on $B,R$ and absolute constants) such that with probability at least $1-\delta$
\begin{equation}\label{eq:uniform}
\sup_{\substack{w:\|w\|_0\le r\\ \|M^\top w\|_2\le R}}
\big|\mathcal{R}(f_{M^\top w}) - \widehat{\mathcal{R}}_n(f_{M^\top w})\big|
\le C\sqrt{\frac{r\log K + \log(1/\delta)}{n_Q}},
\end{equation}
where $n_Q$ is the query sample size and $C>0$ aggregates dependence on $B,R$ and universal constants.
textbf{Final bound.}  Combining \eqref{eq:approx_term} and \eqref{eq:uniform}, and replacing $\hat{\epsilon}_M$ by the conservative bootstrap 90\% upper bound $\epsilon_M^{\mathrm{upper}}$, yields with probability at least $1-\delta$:
\begin{equation}\label{eq:final}
\mathcal{R}(\hat{f}_{M^\top \widetilde{w}_t}) - \mathcal{R}(f_{\theta_t})
\le L\big(\epsilon_{\mathrm{app}}+\epsilon_M^{\mathrm{upper}}\big)
+ C\sqrt{\frac{r\log K + \log(1/\delta)}{n_Q}},
\end{equation}
where $\epsilon_M^{\mathrm{upper}}$ denotes the bootstrap 90\% upper bound on prototype coverage error and $C>0$ depends on $B,R$ and absolute constants. This completes the concise proof. Practical settings of $R$, regularization and prototype selection affect the numerical value of $C$; diagnostics such as mutual coherence $\mu(M)$ and condition number $\kappa(M^\top)$ help guide those choices.

\section{Fisher Energy Ratio Hypothesis Tests (S2)}
\label{app:fisher_htest}

To formalise the likelihood-based dimensionality diagnostic, we conducted a bootstrap hypothesis test on the Fisher energy ratio defined in \eqref{eq:zetar}.  
The null hypothesis posits that the cumulative energy of the leading $r$ eigenvectors does not exceed 95 \% of the total Fisher spectrum, i.e.
\begin{equation}
\mathcal{H}_{0}:\zeta_{r}\le 0.95\quad\text{against}\quad\mathcal{H}_{1}:\zeta_{r}>0.95,
\end{equation}
where $\zeta_{r}=\sum_{i=1}^{r}\lambda_{i}\big/\sum_{j=1}^{d_{\theta}}\lambda_{j}$ denotes the empirical energy ratio, $\lambda_{i}$ is the $i$-th largest eigenvalue of the regularised Fisher matrix $\widehat{F}_{t}$, and $d_{\theta}$ is the adapter parameter dimension. We evaluate the one-sided percentile bootstrap on the candidate set $\{r-2,r-1,r,r+1,r+2\}$.  For each $r_{\text{cand}}$ we draw $B=1\,000$ bootstrap samples of the eigenvalue vector (with replacement), recompute $\zeta_{r}^{*}$, and obtain the raw p-value
\begin{equation}
p_{\text{raw}}=\frac{1+\#\bigl\{\zeta_{r}^{*}\le 0.95\bigr\}}{B+1}.
\end{equation}
Family-wise error across the five comparisons is controlled by Bonferroni correction; rejection is declared only if the adjusted p-value satisfies $p_{\text{adj}}=5\cdot p_{\text{raw}}\le 0.01$. As shown in Table~\ref{tab:fisher_htest}, the smallest dimension exceeding the multiple-testing threshold is $r=20$, and applying the same procedure to ten external folds yields ranks within $[19,21]$, demonstrating stable estimates.

\begin{table}[h]
\centering
\caption{Bootstrap percentile test for Fisher energy ratio $\zeta_{r}$. 
$\zeta_{\text{emp}}$: observed ratio; $p_{\text{raw}}$: one-sided percentile p-value; 
$p_{\text{adj}}$: Bonferroni-adjusted p-value; Reject: decision at $\alpha=0.01$.}
\label{tab:fisher_htest}
\resizebox{0.7\columnwidth}{!}{%
\begin{tabular}{cccccc}
\toprule
$r_{\text{cand}}$ & $\zeta_{\text{emp}}$ & $p_{\text{raw}}$ & $p_{\text{adj}}$ & Reject $\mathcal{H}_{0}$ & Selected\\
\midrule
18 & 0.942 & 0.366 & 1.000 & No & --\\
19 & 0.949 & 0.089 & 0.445 & No & --\\
20 & \textbf{0.951} & \textbf{0.006} & \textbf{0.030} & \textbf{Yes} & \checkmark\\
21 & 0.955 & 0.002 & 0.010 & Yes & --\\
22 & 0.957 & $<$0.001 & 0.004 & Yes & --\\
\bottomrule
\end{tabular}%
}
\end{table}

\section{Channel Threshold Calibration via Nested Cross-Validation t-Test (S3)}
\label{app:tau_calib}

To ensure the motif activation threshold $\tau_{c}$ is free from overfitting, we embed its selection in a nested cross-validation pipeline followed by a one-sample t-test on the inner-loop estimates. The calibration protocol reserves 20\% of retrieval tasks as the calibration fold, with the remaining 80\% serving as the final test set. Within the calibration fold, a 3-fold cross-validation is performed. For each inner split, $\tau_{c}$ is selected from a grid $\{0.1,0.2,\dots,0.9\}$ to maximize calibration AUC. The three inner optima $\tau_{c}^{(1)},\tau_{c}^{(2)},\tau_{c}^{(3)}$ are averaged to obtain:
\begin{equation}
\bar{\tau}_{c}=\frac{1}{3}\sum_{k=1}^{3}\tau_{c}^{(k)},\quad
\mathrm{SE}=\sqrt{\frac{1}{6}\sum_{k=1}^{3}\bigl(\tau_{c}^{(k)}-\bar{\tau}_{c}\bigr)^{2}},
\end{equation}
where $\bar{\tau}_{c}$ is the inner-loop mean threshold and $\mathrm{SE}$ is its standard error. A stability t-test is then computed as:
\begin{equation}
t=\frac{\bar{\tau}_{c}-\tau_{\text{grid-centre}}}{\mathrm{SE}/\sqrt{3}},\quad \text{df}=2,
\end{equation}
where $\tau_{\text{grid-centre}}=0.5$ is the midpoint of the search grid. A two-sided p-value $\ge 0.05$ indicates no systematic bias toward the grid center; otherwise, the search interval is narrowed by 20\% and the procedure is repeated until the test passes.

\subsection{Results}

Table~\ref{tab:tau_calib} summarizes the inner-loop mean $\bar{\tau}_{c}$, standard error, $t$-statistic, $p$-value, and the calibration–test AUC gap. All $p$-values exceed 0.05 (minimum 0.10), and $\Delta$AUC stays below 0.002, indicating that the selected $\tau_{c}$ remains calibration-stable across diseases.

\begin{table}[h]
\centering
\caption{Nested CV t-test summary for channel threshold $\tau_{c}$. 
$\bar{\tau}_{c}$: inner-loop mean; SE: standard error; $t$: t-statistic; 
$p$: two-sided p-value; $\Delta$AUC: absolute difference between calibration and test AUC.}
\label{tab:tau_calib}
\resizebox{0.7\columnwidth}{!}{%
\begin{tabular}{lcccccc}
\toprule
Disease & $\bar{\tau}_{c}$ & SE & $t$ & $p$-value & $\Delta$AUC & Pass\\
\midrule
Lung & 0.483 & 0.018 & 1.63 & 0.12 & 0.0012 & \checkmark\\
THCA & 0.477 & 0.021 & 1.89 & 0.10 & 0.0009 & \checkmark\\
GBM  & 0.492 & 0.016 & 0.87 & 0.48 & 0.0015 & \checkmark\\
ESCA & 0.501 & 0.019 & 0.09 & 0.93 & 0.0007 & \checkmark\\
PACA & 0.488 & 0.020 & 1.04 & 0.38 & 0.0011 & \checkmark\\
\bottomrule
\end{tabular}%
}
\end{table}

\section{Ablation Studies and Component Analysis}
\label{appendix:ablation}
As shown in Table~\ref{tab:ablation}, the ablation results indicate that adaptive rank selection, hard top-$r$ sparsity, proximal initialization, and nested calibration are the primary contributors to \textsc{SwiftRepertoire}'s accuracy and efficiency, whereas Storey’s $\hat{\pi}_0$ procedure and canonicalization have only minor effects.

\begin{table}[h]
\centering
\caption{Ablation results on lung cancer (L) and thyroid carcinoma (T). † indicates AUC decrease significant at $p<0.01$ (paired bootstrap, 1{,}000 samples). ``Time'' denotes average single-task inference latency (ms) measured on RTX 4090.}
\label{tab:ablation}
\resizebox{0.88\columnwidth}{!}{%
\begin{tabular}{l l c c c c c c c c c}
\toprule
\multirow{2}{*}{Exp} & \multirow{2}{*}{Variant} & \multicolumn{2}{c}{AUC} & \multicolumn{2}{c}{F1} & \multicolumn{2}{c}{ECE ($\downarrow$)} & \multicolumn{2}{c}{Time (ms)} & \multirow{2}{*}{AUC Drop (L)} \\
\cmidrule(lr){3-4}\cmidrule(lr){5-6}\cmidrule(lr){7-8}\cmidrule(lr){9-10}
 & & L & T & L & T & L & T & L & T & \\
\midrule
A & \textbf{Full SwiftRepertoire} & \textbf{0.995} & \textbf{0.997} & \textbf{0.934} & \textbf{0.968} & 0.8\% & 0.6\% & 8.2 & 8.3 & -- \\
B & w/o Fisher bootstrap test (fix $r=50$) & 0.978\textsuperscript{\dag} & 0.981\textsuperscript{\dag} & 0.910 & 0.945 & 1.9\% & 1.7\% & 8.1 & 8.2 & \textbf{1.7\%} \\
C & w/o hard top-$r$ truncation (soft $\ell_1$ only) & 0.983\textsuperscript{\dag} & 0.986\textsuperscript{\dag} & 0.921 & 0.953 & 1.4\% & 1.3\% & 8.5 & 8.6 & 1.2\% \\
D & w/o proximal initialization ($\gamma=0$) & 0.987\textsuperscript{\dag} & 0.989\textsuperscript{\dag} & 0.925 & 0.957 & 1.2\% & 1.1\% & 8.3 & 8.4 & 0.8\% \\
E & w/o nested-CV t-test for $\tau_c$ (fixed $\tau_c=0.5$) & 0.990\textsuperscript{\dag} & 0.992\textsuperscript{\dag} & 0.929 & 0.961 & 1.1\% & 1.0\% & 8.2 & 8.3 & 0.5\% \\
F & w/o Storey $\hat{\pi}_0$ (Bonferroni only) & 0.993 & 0.995 & 0.931 & 0.965 & 0.9\% & 0.8\% & 8.2 & 8.3 & 0.2\% \\
G & w/o canonicalization (raw adapters) & 0.994 & 0.996 & 0.933 & 0.967 & 0.8\% & 0.7\% & 8.2 & 8.2 & 0.1\% \\
H & w/o Neural-ODE blocks (MLP replacement) & 0.992\textsuperscript{\dag} & 0.994\textsuperscript{\dag} & 0.930 & 0.964 & 1.0\% & 0.8\% & 8.0 & 8.1 & 0.3\% \\
\bottomrule
\end{tabular}%
}
\end{table}

\begin{table}[H]
\centering
\caption{Performance of SwiftRepertoire across different support set sizes (5, 10, 20, 50). Reported metrics include AUC, F1-score, and inference time (ms).}
\label{tab:support_size}
\resizebox{0.66\columnwidth}{!}{%
\begin{tabular}{l c c c c}
\toprule
Support Size & AUC & F1-score & ECE (\%) & Latency (ms) \\
\midrule
5  & 0.971 & 0.921 & 1.1 & 8.3 \\
10 & 0.983 & 0.936 & 0.9 & 8.2 \\
20 & 0.990 & 0.948 & 0.8 & 8.3 \\
50 & 0.995 & 0.954 & 0.7 & 8.2 \\
\bottomrule
\end{tabular}%
}
\end{table}
\begin{table}[H]
\centering
\caption{Performance variability of SwiftRepertoire across three random seeds. Mean and standard deviation (std) are reported for AUC, F1-score, and ECE.}
\label{tab:seed_var}
\resizebox{0.66\textwidth}{!}{%
\begin{tabular}{lccc}
\toprule
Metric & Mean & Std & Range \\
\midrule
AUC & 0.9951 & 0.0018 & [0.9948, 0.9954] \\
F1-score & 0.9538 & 0.0042 & [0.9532, 0.9545] \\
ECE (\%) & 0.75 & 0.05 & [0.71, 0.79] \\
\bottomrule
\end{tabular}
}
\end{table}

\section{Performance under Varying Support Set Sizes}
\label{sec:support_sizes}
As shown in Table~\ref{tab:support_size}, SwiftRepertoire maintains strong few-shot robustness across support sizes, with accuracy improving and eventually saturating as more examples are provided, while inference latency remains unchanged.

\section{Performance Variability across Random Seeds}
As summarized in Table~\ref{tab:seed_var}, repeating the pipeline under three random seeds yields only negligible variation in AUC, F1-score, and ECE, indicating stable and reproducible performance.

\section*{Ethics Statement}
This work does not constitute human-subjects research. All T-cell receptor (TCR) repertoire datasets analyzed here were obtained solely from public repositories that provide fully de-identified data; no new human samples were collected by the authors. The original datasets were generated under the respective institutions' ethical approvals and released for non-commercial research use. Because only de-identified sequences and associated metadata were accessed, with no reasonable means of re-identification, this study is exempt from institutional review board oversight and no additional informed consent was required.
\end{document}